\documentclass{article}

\usepackage[final, nonatbib]{neurips_2021}

\usepackage{amsmath}
\usepackage[pdftex]{graphicx}

\newcommand{\R}{\mathbb{R}}

\newcommand{\br}[1] {\left( #1 \right)}
\newcommand{\sbr}[1] {\left[ #1 \right]}
\newcommand{\set}[1] {\left\{ #1 \right\}}

\newcommand{\VBP}{{\sf{VBP}}}
\newcommand{\GBP}{{\sf{GBP}}}

\newcommand{\xx}{\boldsymbol{x}}

\usepackage[utf8]{inputenc} 
\usepackage[T1]{fontenc}    
\usepackage{hyperref}       
\usepackage{url}            
\usepackage{booktabs}       
\usepackage{amsfonts}       
\usepackage{nicefrac}       
\usepackage{microtype}      
\usepackage{xcolor}         

\title{Revisiting Sanity Checks for Saliency Maps}

\author{%
  Gal Yona \\
  Weizmann Institute of Science\\
  \texttt{gal.yona@gmail.com} \\
   \And
   Daniel Greenfeld \\
  Jether Energy Research \\
  \texttt{danielgreenfeld3@gmail.com} \\
}

\begin{document}

\maketitle

\begin{abstract}
Saliency methods are a popular approach for model debugging and explainability. However, in the absence of ground-truth data for what the correct maps should be, evaluating and comparing different approaches remains a long-standing challenge.
The sanity checks methodology of  Adebayo \emph{et al} [Neurips 2018] has sought to address this challenge. They argue that some popular saliency methods should not be used for explainability purposes since the maps they produce are not sensitive to the underlying model that is to be explained. Through a causal re-framing of their objective, we argue that their empirical evaluation does not fully establish these conclusions, due to a form of confounding introduced by the tasks they evaluate on. Through various experiments on simple custom tasks we
demonstrate that some of their conclusions may indeed be artifacts of the tasks more than a criticism of the saliency methods themselves. More broadly, our work challenges the utility of the sanity check methodology, and further highlights that saliency map evaluation beyond ad-hoc visual examination remains a fundamental challenge.


\end{abstract}

\section{Introduction}

Feature attribution methods are designed to highlight the features in the input that are important for a given model's predictions. This can be useful for both interpretability and debugging, as a human expert can probe these attributions and verify the extent to which they align with their own priors. In this work, we focus on feature attributions methods in the domain of computer vision, where they are broadly referred to as \emph{saliency maps}. Perhaps the most straightforward approach is to compute the saliency map by deriving the gradient of the model's predicted class w.r.t the input image \cite{simonyan2013deep, erhan2009visualizing}. We refer to this approach as \emph{vanilla backpropogation} (VBP). In recent years, a wealth of different variations of this basic method have been proposed \cite{sundararajan2017axiomatic, smilkov2017smoothgrad}; for example, \emph{guided backpropagation} (GBP) \cite{springenberg2014striving} modifies the gradient computation such that the gradients of negative RELUs are zeroed out. While saliency maps produced using GBP look visually more appealing than those produced with VBP, there is no clear sense in which the former are indeed ``better'' the latter. Indeed, evaluating performance of the different proposed methods (as well as measuring progress) remains a fundamental challenge since in general, the task of saliency maps inherently lacks ``ground truth'' \cite{samek2016evaluating, montavon2018methods, dabkowski2017real, fel2020good, ding2021evaluating, zhou2021feature, arras2020ground, tomsett2020sanity, arun2020assessing}.

In recent work, Adebayo \emph{et al.} \cite{adebayo2018sanity} addressed this challenge by proposing several basic ``sanity checks'' for saliency methods. The simple idea behind the derivation of their tests is that
in so far model debugging and explainability is the objective, using saliency methods that are not sensitive to the underlying model is a bad choice. A classic example are edge detection techniques: while they may produce visually-appealing saliency maps, they are useless for model debugging as the edges depend only on the specific image in question. 
To test whether a particular saliency method $S$ is sensitive to the underlying model, they compare the saliency map on an image $x$ with a saliency map on the same image, with the exception that the underlying model is replaced: e.g., instead of the model $M$, a perturbed model is used (say, $M'$, a randomly initialized network of the same architecture)\footnote{The authors of \cite{adebayo2018sanity} consider two types of procedures for generating the benchmark model $M'$: model randomization (in which the weights of trained model are re-initialized to produce $M'$) and data randomization (in which $M'$ is obtained by training on a perturbed dataset in which the labels of the images have been randomly shuffled). Our focus is on the first.}. If the method depends on the model in question, one would expect the two maps $S(x; M)$ and $S(x; M')$ to be \emph{different}; hence, if they are observed to be \emph{similar}, then we conclude that the method is not sensitive to the model in question. 

\paragraph{This work: a critical perspective.} In their work, 
 \cite{adebayo2018sanity} apply this methodology to  standard pre-trained models on MNIST and ImageNet. Their results are interesting as they suggest an ability to distinguish between the abilities of seemingly similar (and otherwise less understood) saliency methods: for example, vanilla backpropagation and guided backpropagation fare very differently under these randomization tests, with the latter ``failing'' the proposed tests \cite{adebayo2018sanity}. 
 
 Our criticism is that making these conclusions based solely on the mentioned tasks is potentially problematic, as the chocie of task acts as a form of selection bias that may confound the results. For a simple illustration, consider MNIST: Let $\xx$ denote a MNIST image, $M$ a model trained on MNIST, and $M'$ a random model of the same architecture. If $M'$ forms a good representation of the input\footnote{There is increasing evidence that speaks to the power of randomly 
initialized networks of modern deep architectures \cite{saxe2011random, ulyanov2018deep}.} then $S(\xx; M')$ could reasonably highlight the digit in $\xx$ -- which also seems to be the ``correct'' map in this case. Thus, we could very well have that 
 $S(x;M) \approx S(x;M')$ - but this  says nothing about the saliency method $S$ itself! Clearly, MNIST is an extreme example in this context, in that each label corresponds to a single object, which is the only thing in the image. However, one could argue that similar effects may take place in ImageNet, in which images also contain only a single object that corresponds to the correct label, and that is aligned and centered.
 
With these concerns in mind, we propose a causal re-framing of the original objective of \cite{adebayo2018sanity}, from which we derive a modified criterion that makes the dependence on the data-generating distribution (which we refer to as a task) explicit.
We proceed to explicitly design new tasks (based off of both MNIST and ImageNet) in which we expect the representations of a random model will differ significantly from those of a trained model. In particular, we consider  tasks in which the original images are mixed with either \emph{partial} objects or \emph{multiple} objects, and in which the objective is to identify the ``pertrubed'' images. In this way, we guarantee that the ``correct'' saliency map (which in this case we know, having designed the task ourselves) will be visually distinct from that produced by any model that is unaware of the labels. As anticipated, we show that for these simple tasks $S(x;M) \neq S(x;M')$ -- \emph{for both VBP and GBP}. 

On a technical level, our findings call into question the fact that a method like GBP is less-suitable for model debugging purposes \cite{adebayo2020debugging, adebayo2018sanity}. On a broader level, our approach highlights that some of the observations of \cite{adebayo2018sanity} may be an artifact of the tasks they evaluated on more than an actual criticism of the saliency methods themselves. This also raises the concern that the utility of the sanity check methodology itself is limited, in that in practice it may serve as too weak a necessary condition. In our perspective, this highlights the urgency of developing alternative approaches to reliably evaluating and comparing different saliency methods. In this context, it seems natural to explore a more task-specific examination: that is, rather than attempting to rule out a method altogether, determine which of several different methods is better for a given task. In this context, the incorporation of crafted semi-synthetic tasks into the evaluation procedure seems like a natural and interesting direction.

\paragraph{Organization.} The rest of this manuscript is organized as follows. We begin with some preliminaries in Section 2. In Section 3 we propose  an interpretation of the methodology of \cite{adebayo2018sanity} from a causal inference perspective, that tries to formalize the idea of eliminating confounding factors when establishing the effect of model choice on saliency output. In Section 4 we define a variety of custom tasks for which we demonstrate that both guided and vanilla backpropagation fare equally well in terms of the modified sanity checks.

\section{Preliminaries}
\label{sec:prelims}

\paragraph{Saliency methods.}

A model is mapping from an image $x \in \R^d$ to a prediction over $k$-classes, which we interpret as a vector $M(x) \in \R^k$. A saliency method is a mapping from images  $x \in \R^d$ to \emph{saliency maps}, which are also images in $\R^d$. Since the method may depend on the model in question, we use $S(x; M) \in \R^d $ to denote the resulting map. Since our focus is on computer vision, and the models in question are typically deep learning models (and hence differentiable), our focus is on gradient-based saliency methods, namely methods in which we use backpropogation to compute the saliency map. Specifically, we restrict our attention to two commonly used methods: vanilla backpropogation and guided backpropogation. The former simply computes the saliency map as the derivative of the predicted class w.r.t the input image: 
$ S^{\VBP}(x;M) = \frac{\partial \max_j M(x)_j}{\partial x}$;  Guided backpropagation $ S^{\GBP}(x;M)$ is a modification that sets negative entries to zero when backproping through RELU units \cite{springenberg2014striving}. Guided back-propogation tends to produce more visually-appealing saliency maps. Our restriction to these two methods is motivated by the fact that despite their similarity, the sanity checks of \cite{adebayo2018sanity} clearly distinguished them: GBP failed their tests, while VBP passed them. 

\paragraph{Causal graphs.}
Causal inference allows us to formalize the effect of hypothetical actions (interventions), such as the effect of setting the model $\textbf{M}$ to be $M$ or $M'$ on the resulting saliency map. In this work we employ the language of causal graphs. A causal graph is a directed a-cyclic graph in which edges are endowed with a causal interpretation; for a full presentation of the ideas we refer the reader to Pearl's primer \cite{pearl2016causal}. The causal effect of an action $X=x$ on an outcome $Y$ is  written using a special \emph{do} operator, $\textbf{Pr}\sbr{Y=y \vert do(X=x)}$.  A fundamental question in causal inference is when can equations that include the do-operator be re-written in terms of observed variables. In what follows we will be using the \emph{back-door} criterion, which we state below. 

\textbf{Back-door criterion} (Definition 3.3.1 in \cite{pearl2016causal}): 
\begin{itemize}
    \item Given an ordered pair of variables $(X, Y)$ in a
directed acyclic graph $G$, a set of variables $Z$ satisfies the backdoor criterion relative to $(X, Y)$
if no node in $Z$ is a descendant of $X$, and $Z$ blocks every path between $X$ and $Y$ that contains
an arrow into $X$.
    \item If $Z$ satisfies the back-door criterion w.r.t $(X,Y)$, then the causal effect of $X$ on $Y$ is given by the formula $\textbf{Pr}\sbr{Y=y \vert do(X=x)} = \sum_z \textbf{Pr}\sbr{Y=y \vert X=x, Z=z}\cdot \textbf{Pr}(Z=z)$.
\end{itemize}


\section{A causal perspective}

In this section we attempt to use the language of causality to re-interpret the sanity check methodology. Recall that the initial objective of \cite{adebayo2018sanity} is to determine whether a particular saliency method is sensitive to the underlying model. In causal terms, this is like asking whether the \emph{conditional causal effect} of the model choice on the output saliency map is zero. We formalize this by setting up a causal graph for the quantities in question: $T$ is a random variable denoting the \emph{task} in question, where a task $T=t$ is a joint distribution over examples and their respective labels; $X$ is a random variable denoting the image the saliency method will be evaluated on; $M$ is a binary random variable, where $M=1$ means we consider the trained model $M_t$ or the untrained (random) model $M'$. Finally, $S$ is a random variable denoting the eventual output of the saliency method. 
Naturally, we have arrows $X \to S$ and $M \to S$ to capture the fact that the output of the saliency method depends on both the image  and the model in question. The arrow $T \to M$ is because the trained model is trained on the task $T=t$. Finally, we include the arrow $T\to X$ because the saliency method will be evaluated on images from the marginal distribution on images defined by $T=t$. 


The causal effect of model choice (a trained vs random model) on the output saliency map is therefore given by $\Lambda(x)  \equiv \textbf{E}\sbr{S \vert X=x, do(M=1)} - \textbf{E}\sbr{S \vert X=x, do(M=0)} $, where the expectation is over any potential randomness in the saliency method $S$. Note, however, that we cannot simply replace the do-operator with standard conditional expectations, since in this graph the nodes $M$ and $S$ are confounded. Specifically, there is a back-door path  $M \leftarrow T \rightarrow X \rightarrow S$. Informally, Pearl's do-calculus instructs us to ``close`` all back-door paths to compute the causal effect of $M$ on $S$; in our case this is done by conditioning on the fork, $T$. Formally, $\set{T}$ satisfies the back-door criterion w.r.t $(M, S)$ (see Section \ref{sec:prelims} for a formal definition) and so the causal effect should be evaluated as follows:

\begin{equation}
\label{eqn:our-test}
       \Lambda(x) = \sum_t \br{S(x; M_t) - S(x; M') } \cdot \textbf{Pr}(T=t \vert X=x)
\end{equation}

where in the above we replaced $\textbf{E} \sbr{S \vert X= x, T = t, M=1}$ (resp., $M=0
$) with $\textbf{E}  \sbr{S \vert X=x, M_t} = S(x; M_t)$  (resp., $S(x; M')$), under the simplifying assumption that the saliency method is deterministic.

Equation (\ref{eqn:our-test}) is similar to the test performed in \cite{adebayo2018sanity}, only that now the dependence on the tasks in question is made explicit. In the particular case in which we take $T$ to correspond to only ``natural'' tasks, we obtain exactly the test that they performed. However, our main point is exactly that there is no reason to restrict our attention to such tasks when attempting to rule out certain saliency methods, since it is precisely those ``natural'' tasks that may suffer from the confounding effect described above. 

\section{Experiments}
\label{sec:exp}

In this section we proceed to argue that on the basis of the revised test of  Equation (\ref{eqn:our-test}), there is no reason to rule out guided backpropogation (GBP) as a saliency method over vanilla backpropogation (VBP). To do so, we show that there exist a variety of distributions over tasks for which the conditional causal effect of Equation (\ref{eqn:our-test}) is clearly non-zero \emph{for both GBP and VBP}. In particular, we show this for both natural images (e.g. a regular MNIST image) and general images. 
Importantly, our constructions utilize seemingly semi-synthetic tasks. These tasks 
are specifically tailored to ensure that the early layer priors of an untrained network will be different from those of a trained network, and specifically employ partial-object (Section 3.1) and multi-object (Section 3.2) images. In all cases the usage of the synthetic tasks means we do have ground truth data for what the ``correct'' saliency maps \emph{should} look like, and we can indeed verify that both methods produce accurate saliency maps. This strengthens our conclusion that the fact that on \emph{some} tasks the saliency maps produced by GBP resemble those of the maps computed w.r.t random models is an artifact of the tasks, more than a criticism of the method itself.

\subsection{Effect on ``natural'' images}
\label{sec:natural}

\begin{figure}[h]
\centering
{\caption{The two columns on the right show the saliency maps produced by guided backpropogation (GBP) for the image in the left column, for the randomly initialized model and for the model that was trained to predict whether the top half of the image was missing. For MNIST images (top row, bottom row), the difference between the two maps is clearly non-zero.  }\label{fig:mnist_tophalf}}
{\includegraphics[width=0.4\linewidth]{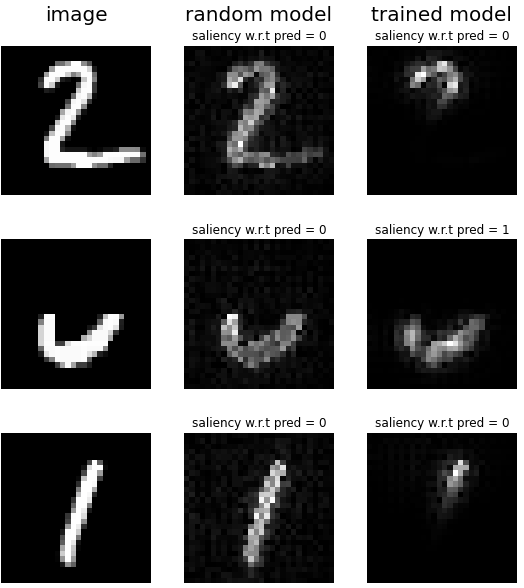}}
\end{figure}

In this section we use $t$ to denote the standard MNIST task., so $M_t$ denotes a standard CNN model trained to near-perfect accuracy on the MNIST dataset and $M'$ denotes a randomly initialized (but untrained) model of the same architecture.
We begin by considering the causal effect of model choice (trained vs random) on the saliency map for images $x \sim t$, i.e., standard MNIST images. We define a simple modification of the standard MNIST task, which we denote $\tilde{t}$, as follows: w.p $1/2$ we perturb the regular MNIST by deleting its bottom half; the objective is to classify images into perturbed and unperturbed images. Naturally this is a simple task; we use $M_{\tilde{t}}$ to denote a model trained to perfect accuracy on this task. We consider the structural causal model in which the distribution of the node $T$ is uniform over the choice of task from $\set{t, \tilde{t}}$. Note that for an (unperturbed) MNIST image $x$, the conditional probability $\Pr(T\vert X=x)$ is non-zero for both tasks; in other words, Equation (\ref{eqn:our-test}) instructs us to consider both the difference w.r.t the regular task $S(x; M_t) - S(x; M')$ and the difference w.r.t the synthetic task $S(x; M_{\tilde{t}}) - S(x; M')$. While for GBP, the first difference is indeed approximately zero \cite{adebayo2018sanity}, the second difference is far from zero, as the results in Figure \ref{fig:mnist_tophalf} demonstrate.

\subsection{Effect on ``synthetic'' images}
\label{sec:non-natural}

In this section we report additional experiments in which we perform the sanity checks w.r.t slightly more complex mluti-object tasks derived from both MNIST and ImageNet. For these tasks we show that the causal effect also on the synthetic images themselves is non-zero. Examining the saliency maps themselves additionally reveals that the outputs of both GBP and VBP are accurate. 

\begin{figure}[h]
\centering
{\caption{The three columns on the right show the saliency maps produced for the different models; from right to left: $M_{\tilde{t}_1}$ (a model trained to predict the shape),  $M_{\tilde{t}_2}$ (a model trained to predict the digit), and a random model of the same architecture.
Through the creation of the multi-object tasks, we see that methods provide  accurate maps (highlighting the object that determines the label) that are visually distinct from the maps of the random model.  }\label{fig:mnist_comp}}
{\includegraphics[width=0.7\linewidth]{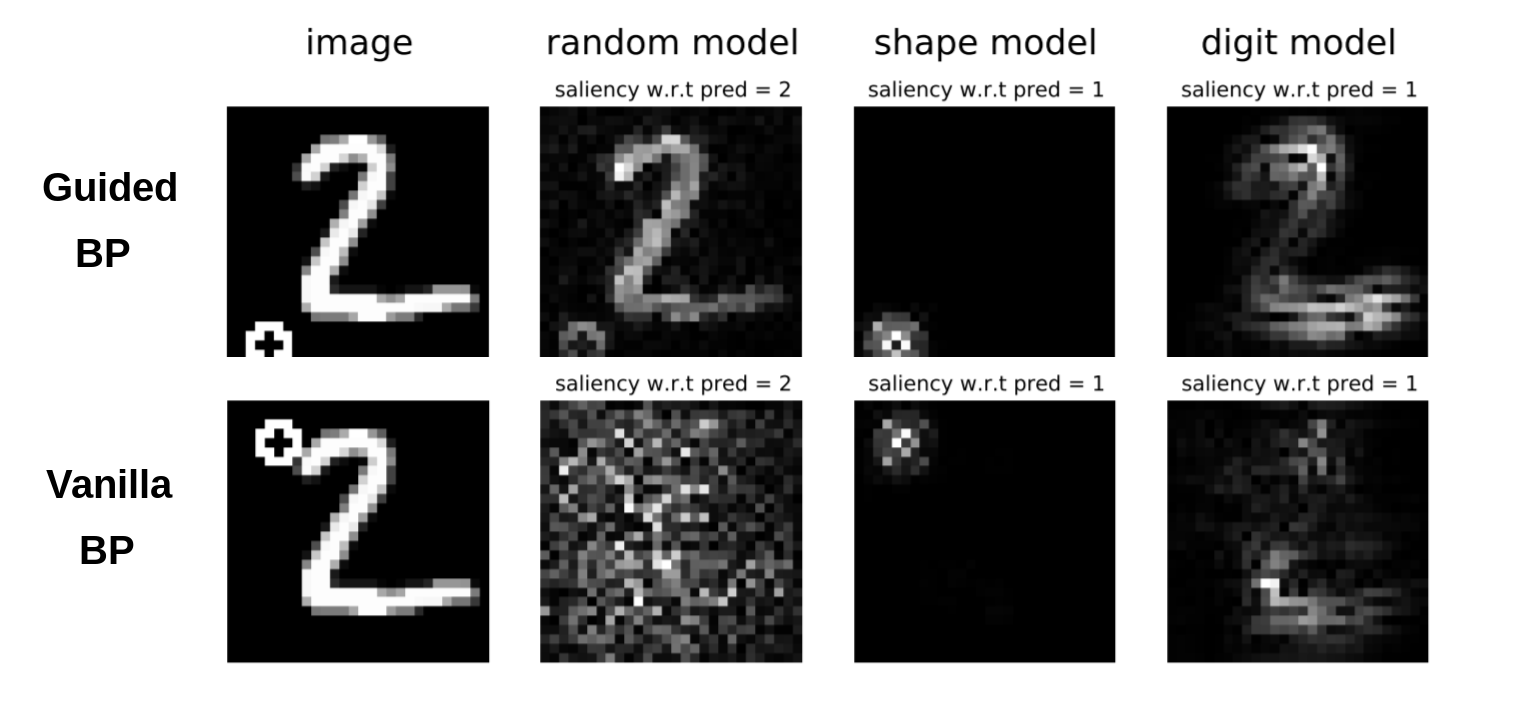}}
\end{figure}

For MNIST, we define a simple modification similar in spirit to the one defined in Section \ref{sec:natural}, only this time we include additional objects in the image. Specifically, images are defined by randomly perturbing MNIST images with one of three modifications (adding a random rectangle, a random empty-circle shape, or doing nothing) and two tasks are defined: in $\tilde{t}_1$ the objective is to determine the modification and in $\tilde{t}_2$ the objective is to determine the digit. We train two models, $M_{\tilde{t}_1}$ and $M_{\tilde{t}_2}$, each to a test accuracy of 0.99 on each task.  Figure \ref{fig:mnist_comp}  shows example saliency maps generated w.r.t each model's predicted label using either GBP (top row) or VBP (bottom row). Clearly, both models produce correct saliency maps that highlight the relevant object in the image\footnote{We note that we only know the ``correct'' maps because in this case we engineered the tasks such that high accuracy is only possible when the model learns to attend to one type of object and ignore the other. In general, as mentioned earlier, we have no knowledge of the ``correct'' saliency maps.} and that are visually distinct from the maps produced w.r.t the random model. See Appendix A for additional examples for each method.

\begin{figure}[htb]
{\caption{The three columns on the right show the saliency maps produced for the different models. From right to left: a ResNet18 model finetuned on the task in question, a pretrained ResNet18 without finetuning, and a random ResNet18. Both methods provide saliency maps that focus only on the relevant image (in this case, the espresso), whereas the other models have positive saliency for both regions of the image, as expected. }\label{fig:imagenet_comp}}
{\includegraphics[width=0.9\linewidth]{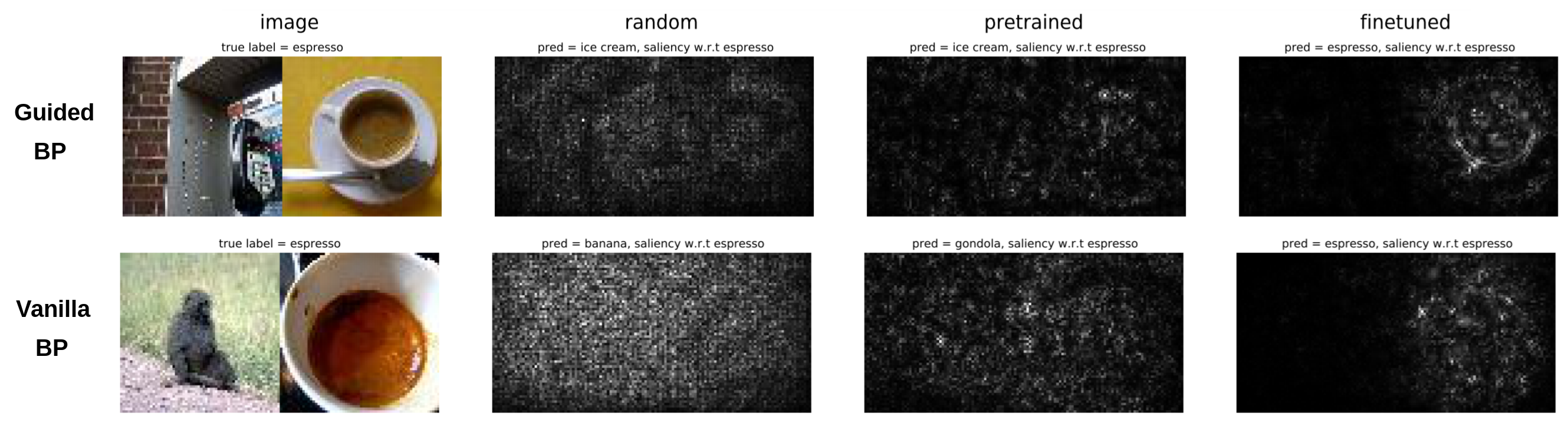}}
\end{figure}

For ImageNet, we generate custom \emph{pairs} of ImageNet images, as follows. First, we consider  the TinyImageNet dataset, which consists of 250 training images and 50 validation images for each original ImageNet class, downscaled to 64x64. We restrict our attention to ten ImageNet classes, and divide them into two groups of five: the first group consists of the labels [acorn, espresso, banana, ice-cream, gondola] and the second consists of the labels [wok, cauliflower, space heater, payphone, baboon]. We then construct a new dataset  as follows: we randomly select one image from the first group (e.g. an espresso) and one image from the second group (e.g. a baboon), where the position is selected randomly. The label of this image pair is defined to be the label of the image that belongs to the first group. We finetune a pretrained ResNet18 model on our pairs dataset until the model reaches a test accuracy of around 0.77. In Figure \ref{fig:imagenet_comp} we demonstrate example saliency maps generated w.r.t the true label using 
either GBP (top row) or VBP (bottom row). We see that 
the two baseline models have positive saliency for both regions of the image -- as expected -- whereas the trained model has positive saliency only on the relevant part of the image (in this case, the espresso, which determines the class label). We refer the reader to Appendix B for additional examples.




\section{Discussion}

In this work we demonstrated empirically that the model randomization tests proposed in \cite{adebayo2018sanity} are  \emph{distribution-dependent}. We have proposed performing the sanity checks w.r.t custom tasks in an attempt to control for factors that may confound the results. Our simple experiments demonstrate that doing so reverses some of the initial observations of \cite{adebayo2018sanity}. We emphasize that this does not challenge the basic idea, which is that for debugging or explainability purposes, one should not use saliency methods which are independent of the model to explained. What it does highlight is that when performed carefully, the necessary condition this perspective proposes may be too weak to provide meaningful distinctions between the existing plethora of saliency methods.  For example, as we have shown in this work, both vanilla backpropogation and guided backpropgation pass the modified sanity checks -- and hence there doesn't seem to be a reason to prefer one over the other on the basis of the sanity check methodology. It would be interesting to explore whether semi-synthetic, engineered tasks, as we used in this work could be used to compare the utility of different methods on a per-task basis.


\clearpage

\bibliographystyle{plain}
\bibliography{refs}

\newpage
\appendix

\section{Additional results for Section \ref{sec:non-natural} (MNIST)}

\begin{figure}[htb]
{\caption{Additional examples: saliency maps produced using vanilla backprop.}}
{\includegraphics[width=0.9\linewidth]{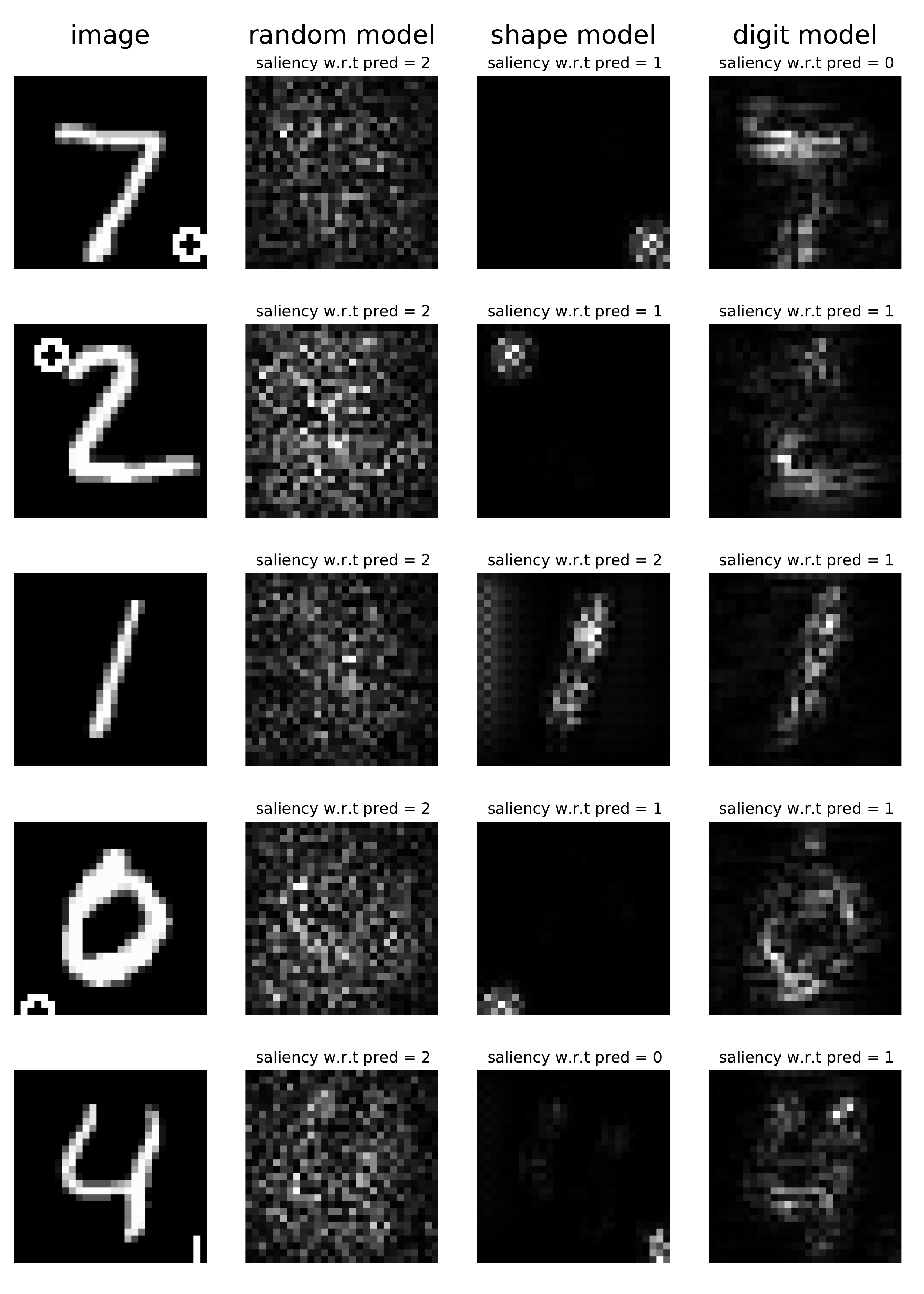}}
\end{figure}

\begin{figure}[htb]
{\caption{Additional examples: saliency maps produced using guided backprop. }}
{\includegraphics[width=0.9\linewidth]{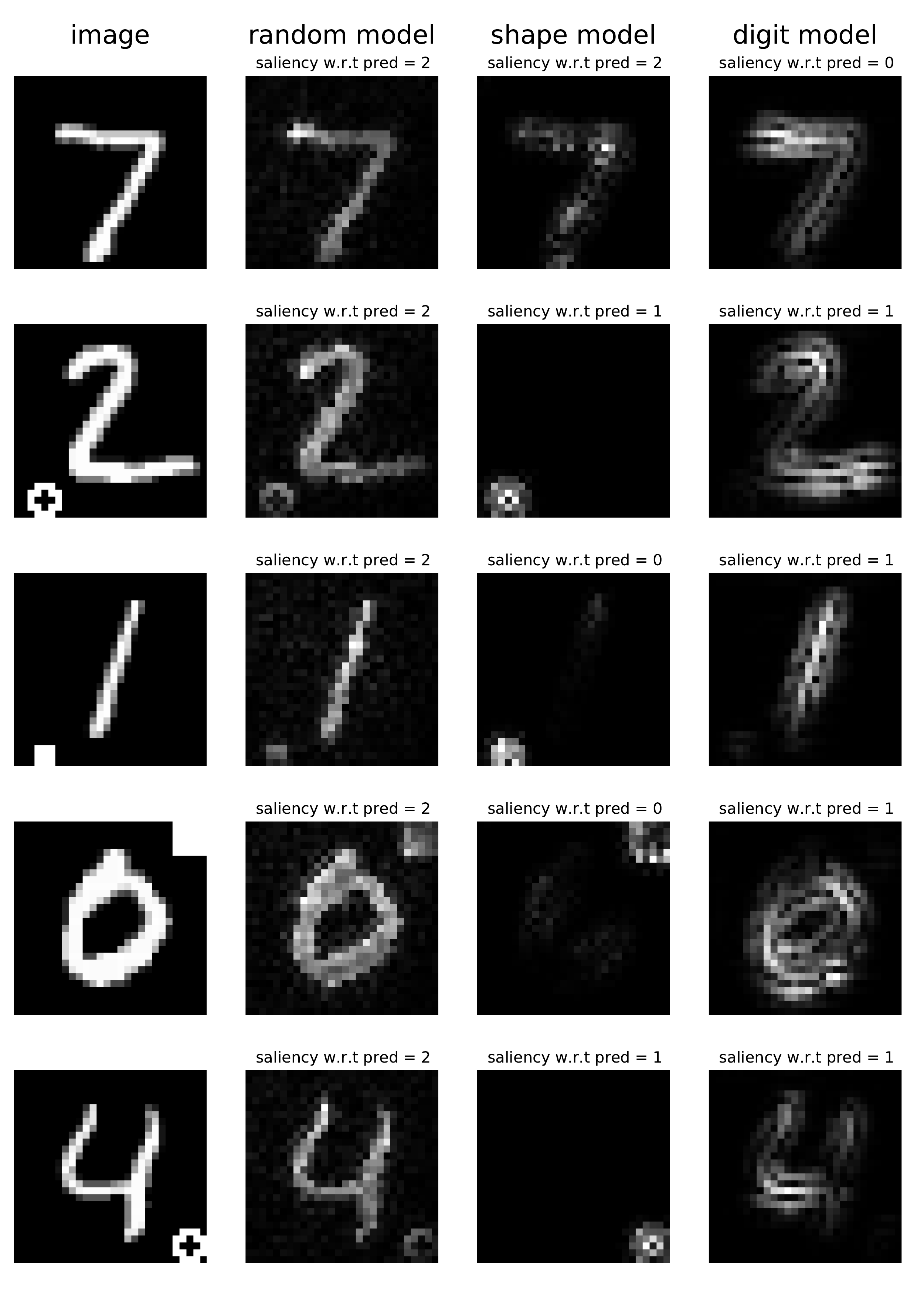}}
\end{figure}

\clearpage
\section{AAdditional results for Section \ref{sec:non-natural} (ImageNet)}

\begin{figure}[htb]
{\caption{Additional examples: saliency maps produced using vanilla backprop.}}
{\includegraphics[width=0.9\linewidth]{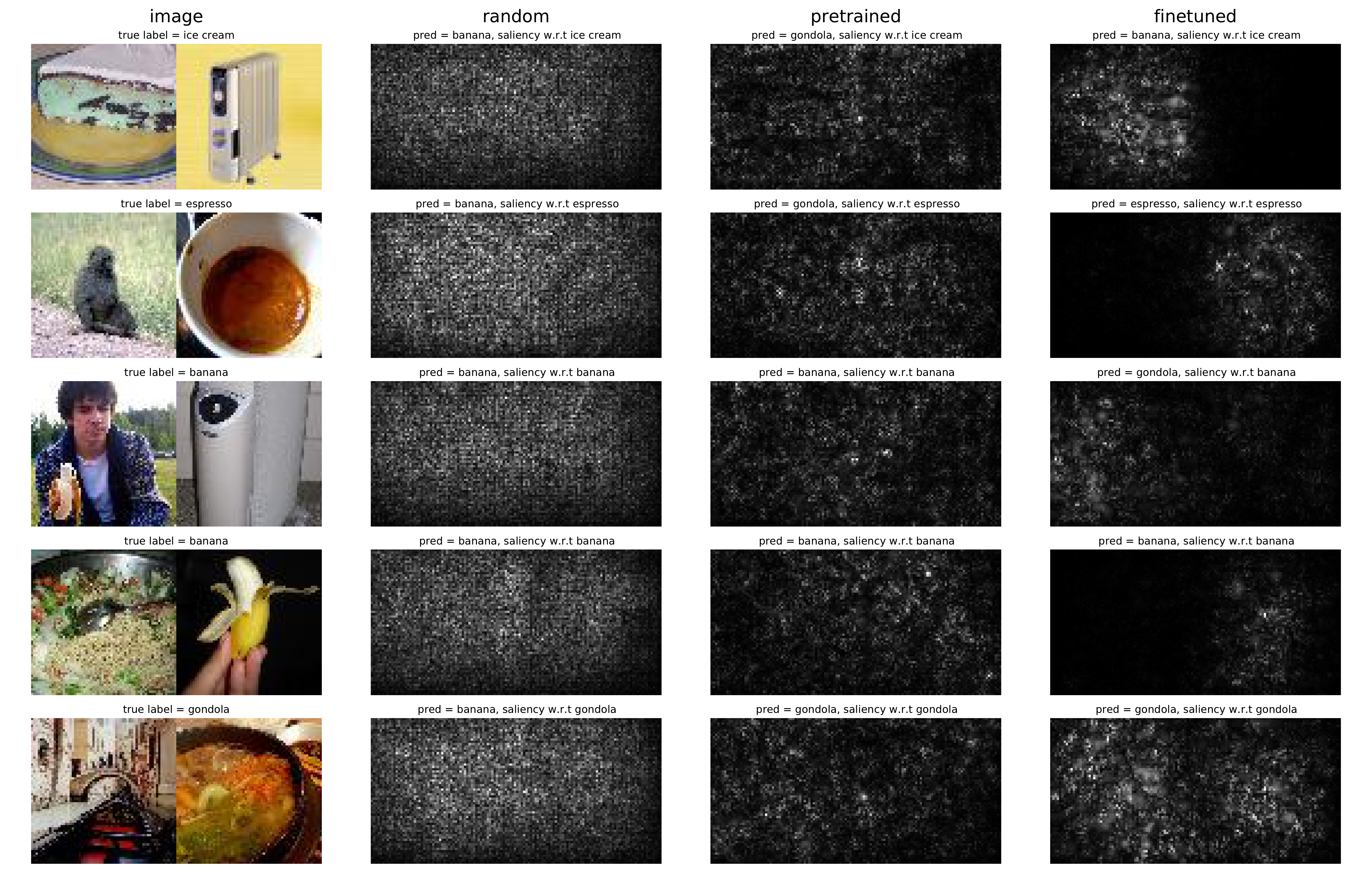}}
\end{figure}

\begin{figure}[htb]
{\caption{Additional examples: saliency maps produced using guided backprop. }}
{\includegraphics[width=0.9\linewidth]{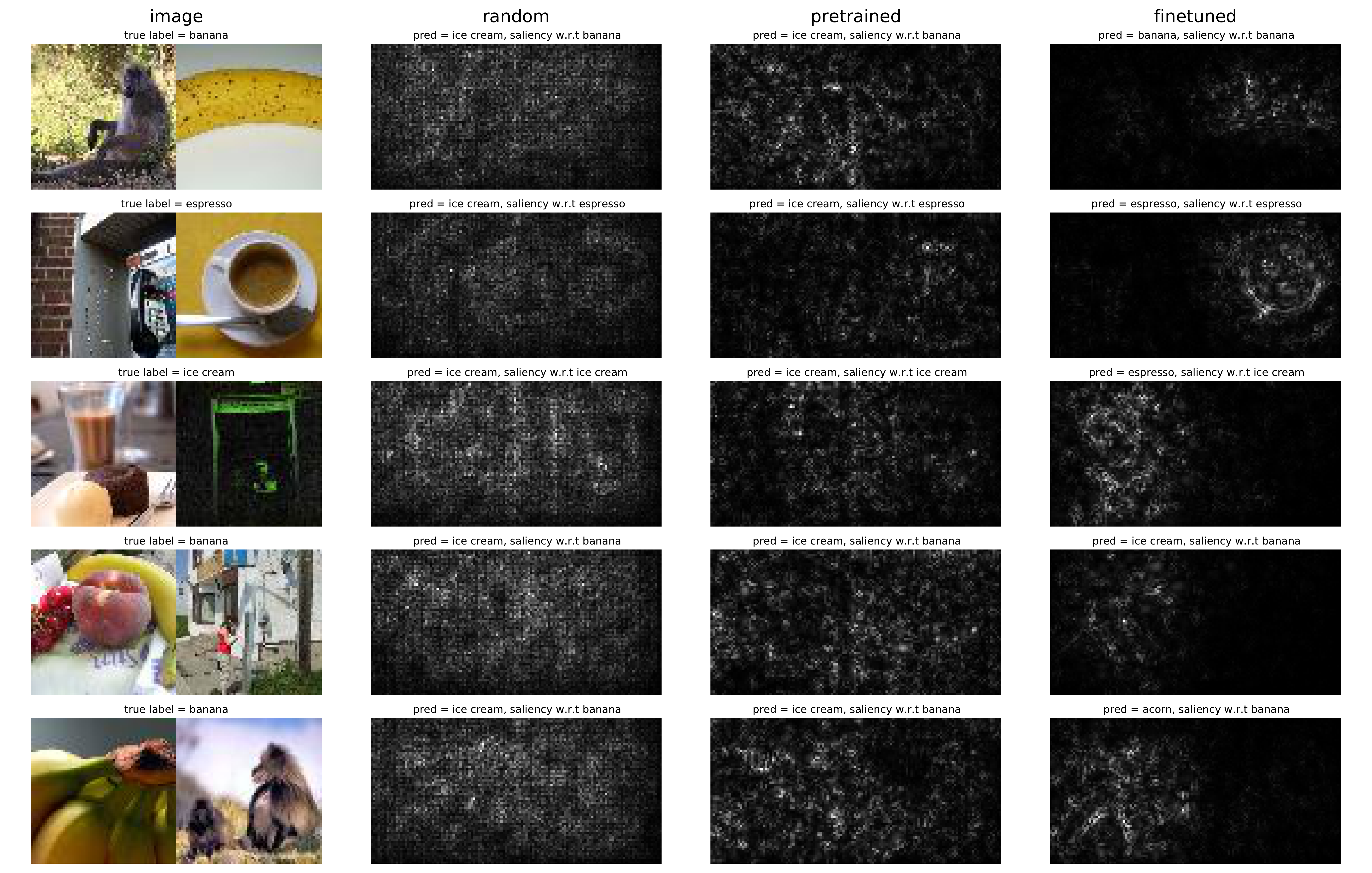}}
\end{figure}

\clearpage

\end{document}